\documentclass{ecai} 

\usepackage{latexsym}
\usepackage{amssymb}
\usepackage{amsmath}
\usepackage{amsthm}
\usepackage{booktabs}
\usepackage{enumitem}
\usepackage{graphicx}
\usepackage{color}

\newcommand{\BibTeX}{B\kern-.05em{\sc i\kern-.025em b}\kern-.08em\TeX}

\begin{document}


\begin{frontmatter}


\paperid{641} 


\title{CLCE: An Approach to Refining Cross-Entropy \\
and Contrastive Learning for Optimized Learning Fusion}


\author[A]{\fnms{Zijun}~\snm{Long}\footnote{Email: longkukuhi@gmail.com}}
\author[A]{\fnms{Lipeng}~\snm{Zhuang}}
\author[A]{\fnms{George}~\snm{Killick}}
\author[A]{\fnms{Zaiqiao}~\snm{Meng}}
\author[A]{\fnms{Richard}~\snm{Mccreadie}}
\author[A]{\fnms{Gerardo}~\snm{Aragon-Camarasa}}

\address[A]{University of Glasgow}


\begin{abstract}
\looseness -1 State-of-the-art pre-trained image models predominantly adopt a two-stage approach: initial unsupervised pre-training on large-scale datasets followed by task-specific fine-tuning using Cross-Entropy loss~(CE). However, it has been demonstrated that CE can compromise model generalization and stability. While recent works employing contrastive learning address some of these limitations by enhancing the quality of embeddings and producing better decision boundaries, they often overlook the importance of hard negative mining and rely on resource intensive and slow training using large sample batches. To counter these issues, we introduce a novel approach named CLCE, which integrates Label-Aware Contrastive Learning with CE. Our approach not only maintains the strengths of both loss functions but also leverages hard negative mining in a synergistic way to enhance performance. Experimental results demonstrate that CLCE significantly outperforms CE in Top-1 accuracy across twelve benchmarks, achieving gains of up to 3.52\% in few-shot learning scenarios and 3.41\% in transfer learning settings with the BEiT-3 model. Importantly, our proposed CLCE approach effectively mitigates the dependency of contrastive learning on large batch sizes such as 4096 samples per batch, a limitation that has previously constrained the application of contrastive learning in budget-limited hardware environments.
\end{abstract}

\end{frontmatter}


\section{Introduction}
\label{sec:intro}

\looseness -1 Approaches for achieving state-of-the-art performance in image classification tasks often employ models initially pre-trained on auxiliary tasks and then fine-tuned on a task-specific labeled dataset with a Cross-Entropy loss (CE)~\citep{RN83, beit3, RN82,RN149,RN150,RN151,RN155}. However, CE's inherent limitations can impact model performance. Specifically, the measure of KL-divergence between one-hot label vectors and model outputs can cause narrow decision margins in the feature space. This hinders generalization~\citep{DBLP:conf/icml/LiuWYY16, DBLP:journals/corr/abs-1906-07413} and has been shown to be sensitive to noisy labels~\citep{DBLP:journals/corr/abs-1901-08360, DBLP:conf/icml/LiuWYY16} or adversarial samples~\citep{RN97, DBLP:journals/corr/abs-1901-08360}. Various techniques have emerged to address these problems, such as knowledge distillation~\citep{DBLP:journals/corr/HintonVD15}, self-training~\citep{DBLP:journals/corr/abs-1905-00546}, Mixup~\citep{DBLP:journals/corr/abs-1710-09412}, CutMix~\citep{DBLP:journals/corr/abs-1905-04899}, and label smoothing~\citep{DBLP:journals/corr/SzegedyVISW15}. However, in scenarios such as few-shot learning, these issues with CE have not been fully mitigated. Indeed, while techniques such as extended fine-tuning epochs and specialized optimizers~\citep{DBLP:journals/corr/abs-2006-05987, DBLP:journals/corr/abs-2006-04884} can reduce the impact of CE to some extent, they introduce new challenges, such as extended training time and increased model complexity~\citep{DBLP:journals/corr/abs-2006-05987, DBLP:journals/corr/abs-2006-04884,RN153}.

Amidst these challenges in context of image classification, contrastive learning has emerged as a promising solution~\citep{RN154}, particularly in few-shot learning scenarios such as CIFAR-FS~\citep{DBLP:conf/iclr/cifarfs} and CUB-200-2011 datasets~\citep{WahCUB_200_2011}. The effectiveness of contrastive learning lies in its ability to amplify similarities among positive pairs (intra-class data points) and distinguish negative pairs (inter-class data points). SimCLR~\citep{RN89}, for instance, has utilized instance-level comparisons unsupervised. However, this unsupervised approach raises concerns regarding its effectiveness , primarily because it limits the positive pairs to be transformed views of an image and treats all other samples in a mini-batch as negatives, potentially overlooking actual positive pairs. We hypothesis incorporating task-specific label information is thus crucial for accurately identifying all positive pairs, especially given the presence of labels in many downstream datasets.

There is a growing trend of using task labels with contrastive learning to replace the standard use of CE~\citep{RN81}. A critical observation here is that many state-of-the-art methods, both in supervised~\citep{RN81,suconlan} and unsupervised~\citep{RN89,npairloss, DBLP:journals/corr/abs-1810-06951,DBLP:conf/ecai/WangLDZB23,DBLP:conf/ecai/WuL23} contrastive learning, overlook the strategic selection of negative samples. They fail to differentiate or prioritize these samples during selection or processing, thereby missing the benefits of leveraging ``hard" negative samples, as highlighted in numerous studies~\citep{DBLP:journals/corr/SongXJS15, DBLP:journals/corr/KumarHC0D17,  suh2019stochastic, DBLP:conf/ijcai/ChuWSJ21, DBLP:conf/ijcai/ZhaoYWYD21, DBLP:conf/ijcai/LiWWZD0Z22}. While contrastive learning mitigates the limitations of CE, it simultaneously introduces a challenge: a reliance on large batch sizes—such as 2048 or 4096 samples per batch—for superior performance compared to CE. This requirement is often impractical in budget hardware environments, particularly when using GPUs with less than 24 GB of memory. As a consequence, state-of-the-art methods such as SupCon~\citep{RN81} underperform compared to CE when using more commonly employed batch sizes, such as 64 or 128 samples per batch, which limits their application. Motivated by these successes and gaps in research, we pose the question: \textit{How can the performance of contrastive learning be improved to address the shortcomings of cross-entropy loss, while also mitigating the reliance on large batch sizes?}

\begin{figure*}
    \centering
    \includegraphics[width=1.0\linewidth]{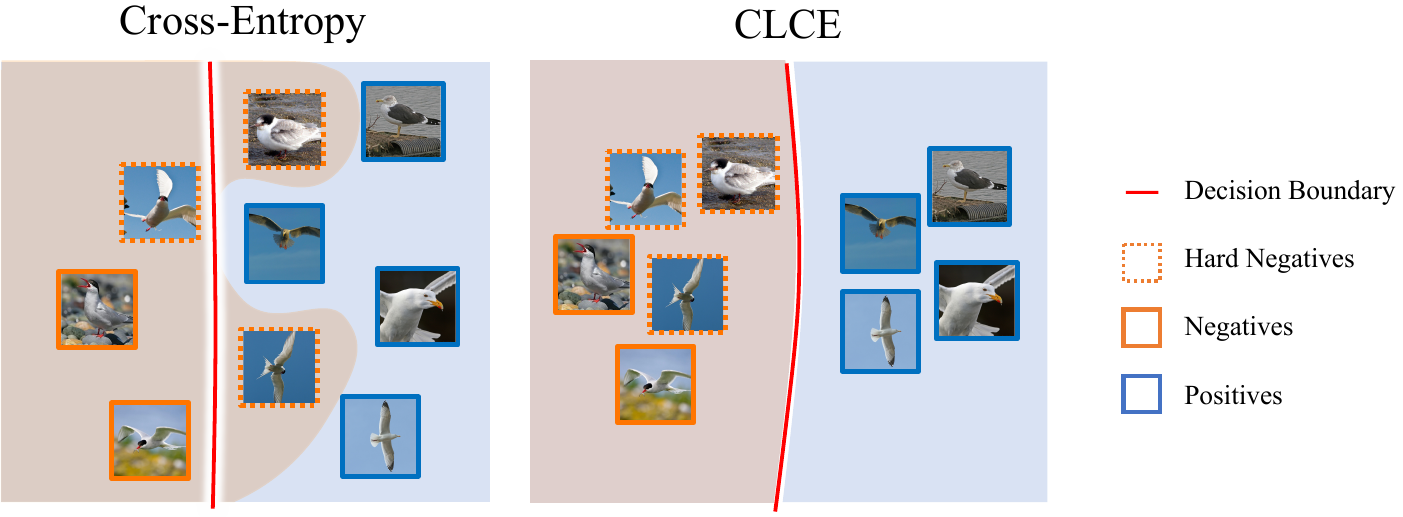}
    \caption{CLCE, our proposed approach, integrates a Label-Aware Contrastive Learning with the Hard Negative Mining (LACLN) term and a CE term. Illustrated with CUB-200-2011 dataset, it emphasizes hard negatives (thick dashed borders) for better class separation. This underscores their marked visual similarity to their positive counterparts. Blue indicates positive examples and orange denotes negatives. On the right, CLCE visibly separates class embeddings more effectively and results a better decision boundary than traditional CE.}
    \label{fig:embeddingvis}
\end{figure*}

Building upon the identified research gaps, we propose CLCE, an innovative approach that combines Label-Aware Contrastive Learning with CE. This approach effectively merges the strengths of both loss functions and integrates hard negative mining. This technique refines the selection of positive and negative samples, thereby enabling CLCE to achieve state-of-the-art performance. As our empirical findings illustrate in Fig.~\ref{fig:embeddingvis}, CLCE places a greater emphasis on hard negative samples that are visually very similar to positive samples, forcing the encoder to learn how to generate more distinct embeddings and better decision boundaries. The core contributions of our work can be summarised as follows:
\begin{itemize}
   \item Introduction of an innovative approach: We introduce CLCE that boosts model performance without necessitating specialized architectures or additional resources. Our work is the first to successfully integrate explicit hard negative mining into Label-Aware Contrastive Learning, retaining the benefits of CE, while also obviating the dependence on large batch sizes.
    \item State-of-the-Art Performance in Few-Shot Learning and Transfer Learning settings: CLCE significantly surpasses CE by an average of 2.74\% in Top-1 accuracy across four few-shot learning datasets when using the BEiT-3 base model~\citep{beit3},  with large gains observed in 1-shot learning scenarios. Additionally, in transfer learning settings, CLCE consistently outperforms other state-of-the-art methods across eight image datasets, setting a new state-of-the-art result for base models (88 million parameters) on ImageNet-1k~\citep{DBLP:conf/cvpr/imagenet}.
    
    \item Reduced Contrastive Learning's Dependency on Large Batch Sizes: Empirical evidence shows that CLCE significantly outperforms both CE and previous state-of-the-art contrastive learning methods like SupCon~\citep{RN81} in commonly used batch sizes, such as 64. This is a size at which earlier state-of-the-art contrastive learning methods underperform. This advancement tackles a crucial bottleneck in contrastive learning, particularly in settings with limited resources. It positions CLCE as a viable, efficient alternative to conventional CE.
\end{itemize}

\section{Related Work}
\subsection{Limitations of Cross-Entropy loss}
The cross-entropy loss (CE) has long been the default setting for many deep neural models due to its ability to optimize classification tasks effectively. However, recent research has revealed several inherent drawbacks~\citep{DBLP:conf/icml/LiuWYY16, DBLP:journals/corr/abs-1906-07413,RN157}. Specifically, models trained with the CE tend to exhibit poor generalization capabilities. This vulnerability stems from the model having narrow decision margins in the feature space, making it more susceptible to errors introduced by noisy labels~\citep{DBLP:journals/corr/abs-1901-08360, DBLP:conf/icml/LiuWYY16} and adversarial examples~\citep{RN97, DBLP:journals/corr/abs-1901-08360}. These deficiencies underscore the need for alternatives that offer better robustness and discrimination capabilities.
\subsection{Contrastive Learning and Negative Mining}
\looseness -1 The exploration of negative samples, particularly hard negatives, in contrastive learning has emerged as a critical yet relatively underexplored area. While the significance of positive sample identification is well-established~\cite{DBLP:conf/ecai/XiaoLC23,ge2021structured,zhang-etal-2024-gla}, recent studies have begun to unravel the intricate role of hard negatives. The potential of hard negative mining in latent spaces has been validated in numerous studies~\cite{suh2019stochastic,DBLP:conf/ijcai/ChuWSJ21,DBLP:conf/ijcai/ZhaoYWYD21,DBLP:conf/ijcai/LiWWZD0Z22,DBLP:conf/ijcai/WuWH22,DBLP:conf/ijcai/YangWJ0PC22,ge2023algrnet,RN152,RN156,,petrov2023gsasrec}. These studies highlight the pivotal role of hard negatives in enhancing the discriminative capability of embeddings. In the contrastive learning domain, \cite{DBLP:journals/corr/abs-2007-00224} tackled the challenge of discerning true negatives from a vast pool of candidates by approximating the true negative distribution. Later, \cite{DBLP:conf/iclr/RobinsonCSJ21} applied hard negative mining to unsupervised contrastive learning, resulting in a framework where only a single positive pair is utilized in each iteration of the loss calculation. However, these approaches still presents limitations, such as inaccurately identifying positive and negative samples and only using one positive pair, which harms the performance of contrastive learning. 
H-SCL \cite{DBLP:journals/corr/abs-2209-00078} expand upon the concept of hard negative mining within a supervised framework. Although both our work and H-SCL utilize hard negative sampling, the methodologies for implementing sampling significantly differ between the two. Their approach employs a consistent threshold-based dot product for identifying ``hard'' samples. However, determining an appropriate threshold remains challenging, as it varies significantly across different datasets and even within individual mini-batches. In contrast, our CLCE method dynamically determines the weighting of each sample, proving to be significantly more effective than H-SCL. Moreover, their methodology does not tackle the dependency on large batch sizes, which is a critical limitation on performance and applicability.

Our work builds on these foundational insights, aiming to synergize the strengths of contrastive learning with CE, particularly by employing hard negative mining guided by label information. CLCE employs a dynamic and adaptive strategy to assign weights to ``hard'' samples in each minibatch, offering a more refined approach compared to previous studies. Additionally, CLCE achieves superior performance to CE without relying on large batch sizes.
\section{APPROACH}\label{sec:Method}

In this paper, we propose an enhanced approach named CLCE for image models that integrates our propose Label-Aware Contrastive Learning with the Hard Negative Mining~(LACLN) and the Cross-Entropy~(CE). CLCE harnesses the potential of contrastive learning to mitigate the limitations inherent in CE while preserving its advantages. Specifically, LACLN enhances similarities between instances of the same class (i.e. positive samples) using label information and contrasts them against instances from other classes (i.e. negative samples), with particular emphasis on hard negative samples. Thus, LACLN reshapes pretrained embeddings into a more distinct and discriminative space, enhancing performance on target tasks. Moreover, CLCE's foundation draws from the premise that the training efficacy of negative samples varies between soft and hard samples. We argue that weighting negative samples based on their dissimilarity to positive samples is more effective than treating them equally. This allows the model to prioritize distinguishing between positive samples and those negative samples that the embedding deems similar to the positive ones, ultimately enhancing overall performance.

\subsection{CLCE}

The overall proposed CLCE approach is a weighted combination of LACLN and standard CE, as expressed in Eq.~\ref{eq:weighted}:

\begin{equation}
    \mathcal{L}_\mathrm{CLCE} = (1-\lambda )\mathcal{L}_\mathrm{CE} + \lambda \mathcal{L}_\mathrm{LACLN}
    \label{eq:weighted}
\end{equation}

In Eq.~\ref{eq:weighted}, the term \(\mathcal{L}_\mathrm{CE}\) represents the CE loss, while \(\mathcal{L}_\mathrm{LACLN}\) symbolizes our proposed LACLN loss. \(\lambda\) represents a scalar weighting hyperparameter. \(\lambda\) determines the relative importance of each of the two losses. 
To provide context for \(\mathcal{L}_\mathrm{CE}\), we refer to the standard definition of the multi-class CE loss, detailed in Eq.~\ref{eq:ce}:

\begin{equation}
\mathcal{L}_\mathrm{CE} = -\frac{1}{N} \sum_{i=1}^{N} \sum_{c=1}^{C} z_{i,c} \log(\hat{z}_{i,c})
\label{eq:ce}
\end{equation}

In Eq.~\ref{eq:ce}, \(z_{i,c}\) and \(\hat{z}_{i,c}\) represent the label and the model's output probability for the \(i\)th instance belonging to class \(c\), respectively.


\begin{figure*}[ht]
\begin{equation}
\resizebox{1\textwidth}{!}{$
\mathcal{L}_{\text{LACLN}} = \displaystyle\sum\limits_{x_{i}\in \mathcal{D}^{*}}- \log \frac{1}{\lvert \mathcal{D}^{*\textbf{+}}_{-x_i}\lvert}  \frac{\displaystyle\sum\limits_{x_{p}\in \mathcal{D}^{*\textbf{+}}_{-x_i}} \exp(x_{i}\cdot x_{p}/\tau)}{\displaystyle\sum\limits_{x_{p}\in \mathcal{D}^{*\textbf{+}}_{-x_i}} \exp(x_{i}\cdot x_{p}/\tau) + \displaystyle\sum\limits_{x_{k}\in \mathcal{D}^{*\textbf{-}}_{-x_i}}   \displaystyle\frac{\lvert \mathcal{D}^{*\textbf{-}}_{-x_i}\lvert}{\displaystyle\sum\limits_{x_{k}\in \mathcal{D}^{*\textbf{-}}_{-x_i}} {\exp(x_{i}\cdot x_{k}/\tau)}} \exp(x_{i}\cdot x_{k}/\tau)^{2}     }
$}
\label{eq:contrastive_loss_fn}
\end{equation}
\end{figure*}

We present the formal definition of our LACLN in Eq.~\ref{eq:contrastive_loss_fn}. This loss introduces a weighting factor for each negative sample, calculated based on the dot product (indicating similarity) between the sample embeddings and the anchor, and normalized by a temperature parameter \(\tau\).  This formulation strategically emphasizes ``hard'' negative samples --- those closely associated with the positive samples by the model's current embeddings. Specifically, the weighting factor for negative samples is determined by calculating their relative proportion based on the average similarity (dot product) observed within each mini-batch. The essence of Eq.~\ref{eq:contrastive_loss_fn} is to minimize the distance between positive pair embeddings and maximize the separation between the anchor and negative samples, particularly the hard negatives. This objective is achieved through two components: the numerator, focusing on bringing positive sample embeddings closer to the anchor, and the denominator, containing both positive and weighted negative samples to ensure the anchor's embedding is distant from negative samples, with a special focus on the more challenging ones. The integration of hard negative mining into contrastive learning is critical as it sharpens the model's ability to differentiate between closely related samples, thus enhancing feature extraction and overall model performance.

Specifically, $\mathcal{D}^{*}$ represents the entire mini-batch composed of an embedding $x$ for each image view (or anchor) $i$. Therefore, $x_i\in\mathcal{D}^{*}$ is a set of embeddings within the mini-batch. The superscripts $+$ and $-$, e.g. $\mathcal{D}^{*\textbf{+}}$, denote sets of embeddings consisting only of positive and negative examples, respectively, for the current anchor within the mini-batch. The term $\lvert \mathcal{D}^{*\textbf{+}}_{-x_i}\lvert$ represents the cardinality of the positive set for the current anchor, while the subscript $-x_i$ denotes that this set excludes the embedding $x_i$. The symbol $\cdot$ represents the dot product. $\tau$ is a scalar temperature parameter controlling class separation. A lower value for $\tau$ encourages the model to differentiate positive and negative instances more distinctly.

\subsection{Analysis of CLCE}

Notably, our proposed CLCE has the following desirable properties:
\begin{itemize}
 \looseness -1 \item Robust Positive/Negative Differentiation: We ensure a clear distinction between true positive and true negative samples by leveraging explicit label information, as encapsulated in Eq.~\ref{eq:contrastive_loss_fn}. This not only prevents the model from being misled by incorrectly contrasting of samples but also reinforces the core philosophy of contrastive learning. The aim is two-fold: to reduce the distance between the embeddings of positive pairs and to increase the distance for negative pairs, ensuring robust class separation.
 \item Discriminating Fine Detail with Hard Negatives: Our loss adjusts the weighting of negative samples based on their similarities to positive instances, as defined in Eq.~\ref{eq:contrastive_loss_fn}. This nuanced approach ensures that the model not only differentiates between glaringly distinct samples but also adeptly distinguishes more challenging, closely related negative samples. Such an approach paves the way for a robust model that discerns real-world scenarios where differences between classes might be minimal.
\end{itemize}

\subsection{Representation Learning Framework}

We use a representation learning framework comprised of three main components, designed specifically to optimize our CLCE approach:

\begin{itemize}
    \item \textbf{Data Augmentation module,} ${Aug(\cdot)}$: This component creates two different views of each sample $r$, denoted $\tilde{r}={Aug(r)}$. This means that every sample will have at least one similar sample (positive pair) in a batch during training.

    \item \textbf{Encoder Network,} ${Enc(\cdot)}$: This network encodes the input data, $r$, into a representation vector, $x=Enc(r)$. Each of the two different views of the data is fed into the encoder separately. 

    \item \textbf{Classification head,} $Head(\cdot)$: This maps the representation vector, \( x \), to probabilities of classes in the target task. The mapping primarily consists of a linear layer, and we utilize its output to calculate the cross-entropy loss.
\end{itemize}

\looseness -1 Our CLCE approach (Eq.~\ref{eq:contrastive_loss_fn}) can be applied using a wide range of encoders, such as BEiT-3~\citep{beit3} or the ResNets~\citep{RN36} for image classification. Following the method in \cite{RN89}, every image in a batch is altered to produce two separate views (anchors). Views with the same label as the anchor are considered positive, while the rest are viewed as negative. The encoder output, represented by $x_{i} = {Enc(r_{i})}$, is used to calculate the contrastive loss. In contrast, the output from the classification head, denoted as $z_{i} = {Head(Enc(r_{i}))}$, is used for the CE. We have incorporated L2 normalization on encoder outputs, a strategy demonstrated to enhance performance significantly~\citep{DBLP:conf/eccv/TianWKTI20}.

\section{Evaluation}\label{sec:eva}
We evaluate our proposed approach, CLCE, on image classification in two settings: few-shot learning and transfer learning. We also conduct several analytical experiments. For CLCE experiments, a grid-based hyperparameter search is conducted on the validation set. Optimal settings (\(\tau=0.5\) and \(\lambda=0.9\)) are employed because they consistently yield the highest validation accuracies. For all experiments, we use the official train/test splits and report the mean Top-1 test accuracy across at least three distinct initializations.

We employ representative models from two categories of architectures -- BEiT-3/MAE/ViT base~\citep{beit3,RN82,RN83} (transformers based models), and ResNet-101~\citep{RN36} (convolutional neural network).
While new state-of-the-art models are continuously emerging (e.g.~DINOv2~\citep{Oquab2023DINOv2LR}), our focus is not on the specific choice of architecture. Instead, we aim to show that CLCE is model-agnostic by demonstrating performance gains with two very different and widely used architectures, as well as show it can be trained and deployed in hardware-constrained settings.
Further implementation details and the complete code for all experiments are publicly available at \url{https://github.com/longkukuhi/CLCE}.



\begin{table*}[]
\vspace{-5mm}
\centering
\resizebox{1\textwidth}{!}{\begin{tabular}{llccccccccc}
\toprule
            &                           & \multicolumn{2}{c}{CIFAR-FS}                                    & \multicolumn{2}{c}{FC100}                                       & \multicolumn{2}{c}{miniImageNet}                                & \multicolumn{2}{c}{tieredImageNet}                              \\ 
Model    & Loss                                 & 1-shot                         & 5-shot                         & 1-shot                         & 5-shot                         & 1-shot                         & 5-shot                         & 1-shot                         & 5-shot                         \\ \midrule

\citep{DBLP:conf/iclr/DhillonCRS20}  & Transductive                    & 76.58$\pm$0.68                     & 85.79$\pm$0.50                     & 43.16$\pm$0.59                     & 57.57$\pm$0.55                     & 65.73$\pm$0.68                     & 78.40$\pm$0.52                     & 73.34$\pm$0.71                     & 85.50$\pm$0.50                     \\
\citep{DBLP:conf/iccv/ZhangMGH21}    & 
Meta-QDA                              & 75.83$\pm$0.88                     & 88.79$\pm$0.75                     & 
-                     & 
-                     & 
67.83$\pm$0.64                     & 84.28$\pm$0.69                     & 74.33$\pm$0.65                     & 89.56$\pm$0.79                     \\ 
\citep{DBLP:conf/nips/HillerMHD22}   & FewTRUE-ViT                                          & \multicolumn{1}{l}{76.10$\pm$0.88} & \multicolumn{1}{l}{86.14$\pm$0.64} & \multicolumn{1}{l}{46.20$\pm$0.79} & \multicolumn{1}{l}{63.14$\pm$0.73} & \multicolumn{1}{l}{68.02$\pm$0.88} & \multicolumn{1}{l}{84.51$\pm$0.53} & \multicolumn{1}{l}{72.96$\pm$0.92} & \multicolumn{1}{l}{87.79$\pm$0.67} \\

\citep{DBLP:conf/nips/HillerMHD22}   
& FewTRUE-Swin                     
& 77.76$\pm$0.81                     & 88.90$\pm$0.59                     & 47.68$\pm$0.78                     & 63.81$\pm$0.75                     & 72.40$\pm$0.78                     & 86.38$\pm$0.49                     & 76.32$\pm$0.87                     & 89.96$\pm$0.55                     \\ 
\citep{DBLP:conf/aistats/0001PG23}    & BAVARDAGE     & 
82.68$\pm$0.25                     & 
89.97$\pm$0.18                     & 
52.60$\pm$0.32                     & 
65.35$\pm$0.25                     & 
77.85$\pm$0.28                     & 
88.02$\pm$0.14                     & 
79.38$\pm$0.29                     & 
88.04$\pm$0.18                     \\ \midrule

ResNet-101         & CE                                       & 69.80$\pm$0.84     & 85.20$\pm$0.62    &43.71 $\pm$0.73   & 58.65$\pm$0.74   & 55.73$\pm$0.85       & 73.86$\pm$0.65      & 46.93$\pm$0.85        & 62.93$\pm$0.76       \\

ResNet-101         & H-SCL~\citep{DBLP:journals/corr/abs-2209-00078}                                       & 67.25$\pm$0.86     & 84.51$\pm$0.65    & 41.34$\pm$0.72   & 57.02$\pm$0.70   & 53.38$\pm$0.79       & 70.29$\pm$0.63      & 44.43$\pm$0.82        & 60.83$\pm$0.71       \\

ResNet-101         & CE+SupCon                                       & 73.61$\pm$0.80     & 86.15$\pm$0.53    & 45.30$\pm$0.62   & 60.18$\pm$0.72   & 57.49$\pm$0.82       & 75.63$\pm$0.61      & 49.44$\pm$0.79        & 66.47$\pm$0.60       \\

ResNet-101         & CLCE (this work)                                       & 76.14$\pm$0.75     & 87.93$\pm$0.48    & 49.48$\pm$0.57   & 64.31$\pm$0.70   & 66.20$\pm$0.74       & 83.41$\pm$0.55      & 63.61$\pm$0.72        & 79.83$\pm$0.51       \\

BEiT-3         & CE                                       & 83.68$\pm$0.80     & 93.01$\pm$0.38    & 66.35$\pm$0.95   & 84.33$\pm$0.54   & 90.62$\pm$0.60       & 95.77$\pm$0.28      & 84.84$\pm$0.70        & 94.81$\pm$0.34       \\

BEiT-3         & H-SCL~\citep{DBLP:journals/corr/abs-2209-00078}                                       & 82.21$\pm$0.80     & 91.49$\pm$0.37    & 65.27$\pm$0.98   & 82.61$\pm$0.52   & 88.57$\pm$0.62       & 93.03$\pm$0.29      & 81.37$\pm$0.73        & 93.26$\pm$0.33       \\

BEiT-3         & CE+SupCon                                       & 84.93$\pm$0.74    & 93.36$\pm$0.34    & 67.58$\pm$0.86   & 86.10$\pm$0.57   & 91.04$\pm$0.55       & 95.97$\pm$0.24      & 85.72$\pm$0.64        & 95.33$\pm$0.29       \\

BEiT-3         & CLCE (this work)                        & \textbf{87.00$\pm$0.70}     & \textbf{93.77$\pm$0.36}    & \textbf{69.87$\pm$0.91}   & \textbf{87.06$\pm$0.52}   & \textbf{92.35$\pm$0.53 }      & \textbf{96.78$\pm$0.23}      & \textbf{87.24$\pm$0.62}        & \textbf{96.09$\pm$0.29}      \\ \bottomrule
\end{tabular}}
\caption{Comparison to baselines on the few-shot learning setting. Average few-shot classification accuracies (\%) with 95\% confidence intervals on test splits of four few-shot learning datasets. }
\label{tab:FS}
\end{table*} 




\begin{table*}[]
\vspace{-3mm}
\centering
\resizebox{0.99\textwidth}{!}{\begin{tabular}{@{}llcccccccc@{}}
\toprule
Model &  Loss   & CIFAR-100      & CUB-200        & Caltech-256     & Oxford-Flowers & Oxford-Pets  & iNat2017       & Places365 & ImageNet-1k               \\  \midrule
ResNet-101 & CE                                       & 96.27         & 84.62          & 81.38          & 95.71          & 93.24 & 66.11 & 54.73 & 78.70      \\

ResNet-101 & H-SCL~\citep{DBLP:journals/corr/abs-2209-00078}                                       &92.78          &77.14           &78.64           &92.34           &92.58  &63.14  &52.02  &77.10       \\

ResNet-101 & CE+SupCon                                       & 96.31   & 84.70           & 81.61          &  95.73    & 93.49 & 66.90 &  55.41  & 79.03      \\

ResNet-101   & CLCE (this work)                                        & \textbf{96.92}          & 87.48          & 85.05          & \textbf{96.33}          & 94.21 & 67.93  & 57.30   &  80.16
\\  \midrule

ViT-B & CE                                       & 87.13          & 76.93          & 90.92          & 90.86          & 93.81 &65.26 &54.06 &77.91                      \\
ViT-B   & CLCE (this work)    & 88.53 & 78.21 &  92.10 & 92.04 & 94.01  & 71.25 & 58.70 &  83.94                                   \\
 \midrule
MAE & CE                                       & 87.67          & 78.46          & 91.82          & 91.67          & 94.05  &70.50 &57.90 &83.60       \\

MAE   & CLCE (this work)                                        & 90.29          & 81.30          & 93.11          & 92.82          & 94.88 & 71.62 & 58.40  & 84.02    \\  \midrule

BEiT-3 & CE                                       & 92.96          & 98.00          & 98.53          & 94.94          & 94.49 &72.31 &59.81 &85.40              \\
BEiT-3 & H-SCL~\citep{DBLP:journals/corr/abs-2209-00078}                                       &89.50          &95.70           &96.24           &92.60           &93.28  &68.51  &56.66  &82.25       \\
BEiT-3 & CE+SupCon & 92.74  & 98.06 & 98.65 & 94.92 & 94.77 & 73.58 & 60.52 & 85.70                     \\
BEiT-3 & CLCE (this work)                        & 93.56 & \textbf{98.93} & \textbf{99.41} & 95.43 & \textbf{95.62} & \textbf{75.72}& \textbf{62.22}& \textbf{86.14} \\ \bottomrule
\end{tabular}}
\caption{Comparison to baselines on transfer learning setting. The results are Top-1 classification accuracies across eight diverse datasets.}
\label{tab:FD}
\vspace{-2mm}
\end{table*}

\subsection{Few-shot Learning}

We evaluate our proposed CLCE in the few-shot learning setting. The experiments on few-shot learning aim to assess the quality of the learned representations.  Specifically, each test run comprises 3,000 randomly sampled tasks, and we report median Top-1 accuracy with a 95\% confidence interval across three runs, maintaining a consistent query shot count of 15. Four prominent benchmarks are used for evaluation: CIFAR-FS~\citep{DBLP:conf/iclr/cifarfs}, FC100~\citep{DBLP:conf/nips/fc100}, miniImageNet~\citep{DBLP:conf/nips/miniimagenet}, and tieredImageNet~\citep{DBLP:conf/iclr/tieredimagenet}. We follow established splitting protocols for a fair comparison~\citep{DBLP:conf/iclr/cifarfs,DBLP:conf/nips/fc100,DBLP:conf/iclr/RaviL17}.

\looseness -1 Tab.~\ref{tab:FS} shows the performance of BEiT-3 and ResNet-101 models under various methods, including CE, H-SCL~\citep{DBLP:journals/corr/abs-2209-00078}, and the same weighted combination of CE and state-of-the-art supervised contrastive learning loss (SupCon)~\citep{RN81} as CLCE. The results reveal that our CLCE approach consistently improves classification accuracy over other methods, demonstrating superior generalization with limited training data for each class.
Our CLCE enhances models' performance on few-shot datasets, significantly outperforming both CE and CE+SuperCon (paired t-test, p $<$ 0.01). In the 1-shot learning context when compared to BEiT-3 trained with CE (BEiT-3-CE), the most remarkable improvement is seen on the FC100 dataset, with accuracy rising by $3.52\%$ through the use of CLCE (BEiT-3-CLCE). Indeed, across all datasets, BEiT-3-CLCE shows an average accuracy improvement of $2.7\%$. For 5-shot learning, the average improvements across the datasets are $1.4\%$ in accuracy for BEiT-3-CLCE, demonstrating CLCE’s effectiveness in scenarios with fewer positive samples per class and its ability to yield consistent and reliable results, evident in the tighter confidence intervals for Top-1 accuracy. As for ResNet-101, CLCE (ResNet-101-CLCE) demonstrates even more significant improvements over both CE and CE+SupCon. The enhancement is especially remarkable in the case of tieredImagenet, where ResNet-101-CLCE achieves increases of $16.68\%$ over ResNet-101-CE and $14.17\%$ over ResNet-101-CE+SupCon in 1-shot learning. For 5-shot learning, the improvements are $16.9\%$ and $13.36\%$, respectively. On average, ResNet-101-CLCE achieves a $9.82\%$ improvement in 1-shot and an $8.71\%$ improvement in 5-shot settings over the ResNet-101-CE. Lastly, H-SCL~\citep{DBLP:journals/corr/abs-2209-00078} underperforms compared to CE at a batch size of 128. This highlights contrastive learning's limitation of needing very large batch sizes for better performance than CE, evident in ResNet-101 and BEiT-3 models.

Overall, the enhancement of our CLCE is particularly effective for few-shot scenarios, where limited labelled data requires the model to rely more on high-quality, discriminative representations. These outcomes underline the efficacy of our proposed CLCE approach and CLCE's broad applicability across different model architectures for few-shot learning tasks.

\subsection{Transfer Learning}
\looseness -1 We now assess the transfer learning performance of our proposed CLCE.
Here, adhering to the widely accepted paradigm for achieving state-of-the-art results, models are initialized with publicly-available weights from pretraining on ImageNet-21k~\citep{DBLP:conf/cvpr/imagenet} since they are state-of-the-art, and are fine-tuned on smaller datasets using our new loss function.
We leverage 8 datasets: CIFAR-100~\citep{krizhevsky2009learning}, CUB-200-2011~\citep{WahCUB_200_2011}, Caltech-256~\citep{caltech256}, Oxford 102 Flowers~\citep{DBLP:conf/icvgip/flowers}, Oxford-IIIT Pets~\citep{DBLP:conf/cvpr/pets}, iNaturalist 2017~\citep{DBLP:conf/cvpr/inat2017}, Places365~\citep{DBLP:conf/nips/places}, and ImageNet-1k~\citep{DBLP:conf/cvpr/imagenet}. We adhere to official train/test splits and report mean Top-1 test accuracy over three different initializations.

Tab.~\ref{tab:FD} presents the results of transfer learning, which offers further evidence of the effectiveness of our proposed CLCE approach beyond few-shot scenarios. When applied to four state-of-the-art image models, including BEiT-3, ResNet-101, ViT-B and MAE, our proposed CLCE approach consistently surpasses other methods, including the standard CE, H-SCL~\citep{DBLP:journals/corr/abs-2209-00078} and the same weighted combination of CE and SupCon loss as CLCE. A paired t-test confirms these improvements as statistically significant ($p<0.05$).
While the increase in performance with BEiT-3-CLCE over the BEiT-3-CE baseline is modest in some cases, such as the rise from $98.00\%$ (BEiT-3-CE) to $98.93\%$ (BEiT-3-CLCE) on CUB-200, it shows significant enhancements in challenging datasets with a higher level of class diversity. A notable example is iNaturalist2017, which has 5089 different classes, where CLCE leads to a marked improvement in accuracy from 72.31\% to 75.72\%. This substantial increase suggests that CLCE's benefits are more pronounced in more varied datasets. In the case of ImageNet-1k, accuracy increased from $85.40\%$ (BEiT-3-CE) to $86.14\%$ (BEiT-3-CLCE), setting a new state-of-the-art for base models (88 million parameters)~\footnote{\url{https://paperswithcode.com/sota/image-classification-on-imagenet}}. We observe similar improvements in other transformer-based models, such as ViT and MAE. The use of CLCE in fine-tuning ResNet-101 also resulted in significant performance gains, particularly in the Caltech-256 dataset. Here, the model's accuracy increases from $81.38\%$ (ResNet-101-CE) to $85.05\%$ (ResNet-101-CLCE). Compared to ResNet-101-CE, there has been an average increase in accuracy of $1.83\%$ for ResNet-101-CLCE. Furthermore, H-SCL~\citep{DBLP:journals/corr/abs-2209-00078} yields inferior results compared to CE, mirroring the result observed in few-shot scenarios.
Overall, the consistent achievement of high accuracies across diverse datasets using models fine-tuned with CLCE, especially ResNet-101 and BEiT-3, underscores the effectiveness of CLCE in improving model performance. Remarkably, this is achieved without resorting to specialized architectures, extra data, or heightened computational requirements, thereby establishing CLCE as a powerful alternative to traditional CE.

\subsection{Reducing Batch Size Dependency}
\looseness -1 We evaluate the effect of batch size on the performance, specifically comparing our CLCE approach with CE and SupCon~\citep{RN81}. The results, as detailed in Tab.~\ref{tab:batchsize}, indicate that SupCon's performance is sensitive to batch size variations, a limitation not observed with CE. Particularly, SupCon shows inferior performance compared to CE with the commonly used batch size of 64 on both tested datasets. Even when the batch size is increased to 128, SupCon continues to underperform relative to CE. In our experiments, SupCon generally needs a batch size exceeding 512 to outperform CE, a requirement that is impractical for most single-GPU setups. This scenario mirrors the results of H-SCL~\citep{DBLP:journals/corr/abs-2209-00078} in the context of few-shot and transfer learning. In contrast, CLCE not only surpasses CE performance on the iNat2017 dataset with a 1.41\% accuracy improvement with batch size of 64 but also demonstrates an even more performance gain of 3.52\% in accuracy with batch size of 128. Thus, our CLCE approach significantly mitigates the dependency on large batch sizes typically associated with contrastive learning approaches like SupCon and H-SCL. The reduction in dependency on large batch sizes greatly enhances the adaptability and effectiveness of CLCE in diverse computational settings, such as environments with budget GPUs equipped with 12 GB of memory.

\looseness -1 Moreover, gradient accumulation is commonly used in cross-entropy loss to achieve a similar effect when requiring large batch sizes. However, gradient accumulation is very challenging in contrastive learning due to the need to ensure that the accumulated gradients accurately reflect the contrastive nature of the task, particularly in maintaining the integrity of positive and negative pair distributions. This also increases the complexity of maintaining effective sampling strategies which could vary among datasets, in pairs or triplets across accumulation steps. Thus, gradient accumulation is an inadequate method for overcoming the dependency on large batch sizes. CLCE, on the other hand, offers a more efficient and effective solution.

 \begin{figure}
     \centering
     \includegraphics[width=0.9\linewidth]{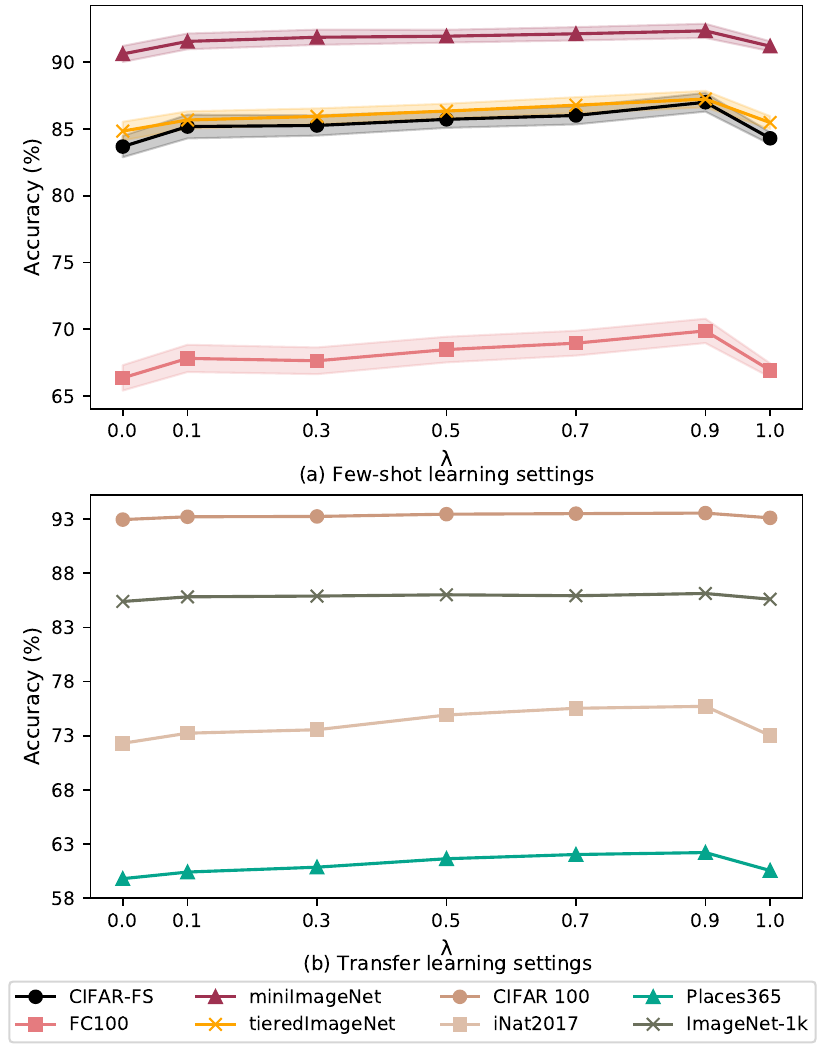}     \caption{Evaluation of the impact of the $\lambda$ hyperparameter. Results on eight tested datasets with $\lambda$ values ranging from $\{0, 0.1, 0.3, 0.5, 0.7, 0.9, 1.0\}$. The numerical details for these figures are provided in the supplementary material \cite{long2024clceapproachrefiningcrossentropy}.}
     \label{fig:lambdasweep}
 \end{figure}

\begin{table}[t]
\centering
\resizebox{8.2cm}{!}{
\tiny
\begin{tabular}{@{}lccc@{}}
\toprule
 Loss  & \multicolumn{1}{l}{Batch Size} & \multicolumn{1}{l}{CIFAR-FS} & \multicolumn{1}{l}{iNat2017} \\ \midrule
 CE                  &  64 & 83.68 & 72.31      \\
CE                  &  128 & 83.39  & 72.20      \\
  \midrule

 SupCon~\citep{RN81}         & 64 & 80.31  & 69.05        \\
 SupCon~\citep{RN81}        & 128 & 82.17  & 69.93
\\ \midrule
 CLCE (this work)        & 64 &   84.59 & 73.72        \\
 CLCE (this work)        & 128 & 87.00   & 75.72     \\ \bottomrule
\end{tabular}}
\caption{\looseness -1 Impact of different batch size. Performance of BEiT-3 base model when trained on CIFAR-FS and iNat2017 datasets. ``CE" denotes cross-entropy loss. ``SupCon" denotes supervised contrastive learning loss. ``CLCE" denotes our proposed joint loss.
}
\label{tab:batchsize}
\end{table}

\subsection{Optimizing \(\lambda\): Bridging CE and LACLN}
\looseness -1 Our proposed CLCE incorporates a hyperparameter, \(\lambda\), to control the contributions of the CE term and the proposed LACLN term, as shown in Eq.~\ref{eq:weighted}. To understand the influence of \(\lambda\), we evaluate its effect on classification accuracy in few-shot learning and transfer learning. Fig.~\ref{fig:lambdasweep} presents the test accuracy for varying values of \(\lambda\). Our experiments reveal a consistent trend: as the weight assigned to the LACLN term (\(\lambda\)) increases, performance progressively improves across all tested datasets, peaking at \(\lambda = 0.9\). For instance, this optimal setting yields an average performance boost of $2.14\%$ and $2.74\%$ over the exclusive use of either the LACLN or CE term on four few-shot datasets.  This trend also manifests in transfer learning settings, highlighting the complementary nature of CE and LACLN. Thus, optimizing this balance is crucial for maximizing performance with CLCE.

\begin{table}[]
\centering
\resizebox{8.2cm}{!}{
\tiny
\begin{tabular}{@{}lcccc@{}}
\toprule
 CE & CL & HNM & CIFAR-FS & iNat2017 \\ \midrule
 \checkmark &  &    & 83.68 & 72.31  \\
 \checkmark     & \checkmark   &   &84.85 &  73.53       \\
 \checkmark        & \checkmark & \checkmark  &87.00 & 75.72      \\ \bottomrule
\end{tabular}
}
\caption{\looseness -1 Results on CIFAR-FS and iNat2017 when training BEiT-3 base model using ablated versions of our CLCE. ``CE" denotes cross-entropy loss. ``CL" refers to our proposed label-aware contrastive learning, and ``HNM" refers to hard negative mining.}
\label{tab:ablation}
\end{table}

\subsection{Ablation Study}
We conducted an ablation study on the CIFAR-FS and iNat2017 datasets to evaluate the contributions of two key components in our proposed loss: the proposed label-aware contrastive learning loss without hard negative mining~(CL), and the proposed hard negative mining strategy~(HNM), as presented in Tab.~\ref{tab:ablation}. Across both tested datasets, integrating CL with CE
is essential for achieving better performance than the CE---e.g.~on the CIFAR-FS dataset, there is a notable performance increase of \(1.17\%\). Meanwhile, the integration of our proposed HNM is critical for CLCE’s enhanced performance, representing one of the main contributions of this paper. For example, it yields a gain of 2.19\% accuracy on the iNat2017 dataset compared to the variant of CLCE without HNM. Hence, we conclude that both components are important and complementary.


\begin{figure}[h]
\centering
\includegraphics[width=1\linewidth]{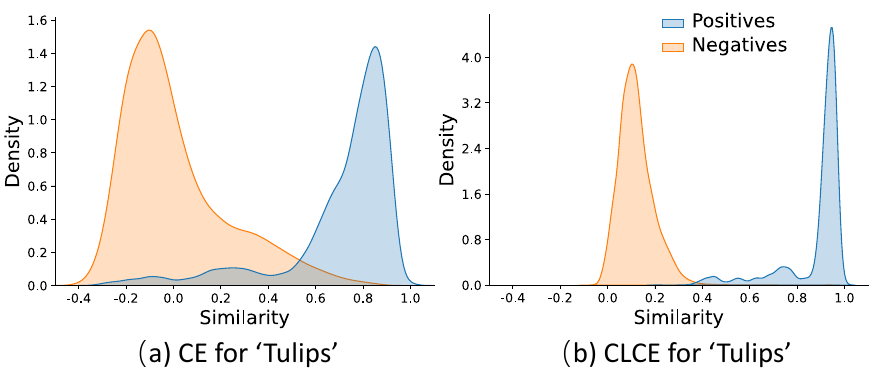}
\caption{Plot of cosine similarity distribution across the ``tulips'' class from CIFAR-100. Blue represents similarities of positive samples, while orange represents similarities of negative samples.}
\label{fig:cosine1}
\end{figure}

\begin{figure}[h]
\centering
\includegraphics[width=0.9\linewidth]{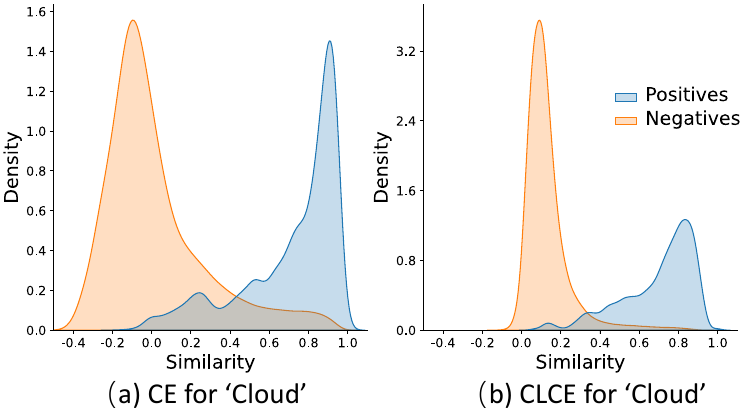}
\caption{Plot of cosine similarity distribution across the ``cloud'' class from CIFAR-100. Blue represents similarities of positive samples, while orange represents similarities of negative samples.}
\label{fig:cosine2}
\end{figure}

\subsection{Embedding Quality Analysis}

To validate the enhancements brought by the proposed approach, CLCE, we perform a thorough evaluation focusing on the geometric characteristics of the generated representation spaces. We hypothesize that our CLCE enhances the quality of embeddings, thereby sharpening class distinction and improving performance. To elaborate, we examine the CE embeddings and CLCE embeddings produced by the BEiT-3 base model. Specifically, we evaluate two key aspects:~(1) Distributions of cosine similarities between image pairs. This assessment provides insights into how well the model differentiates between classes in the embedding space. (2) Visualization of the embedding space using the t-SNE algorithm~\citep{tsne}. This visualization allows us to observe the separation or clustering of data points belonging to different classes. (3) We employ the Isotropy Score as defined by \cite{mu2018allbutthetop} to evaluate the quality of produced embeddings. The Isotropy Score measures the distribution of data in the embedding space and serves as a metric for the quality of the produced embeddings. Historically, isotropy has served as an evaluation metric for representation quality~\citep{arora2016latent}. This is based on the premise that widely distributed representations across different classes in the embedding space facilitate better distinction between them.

We present the pairwise cosine similarity distributions of CE and CLCE embeddings in Figs.~\ref{fig:cosine1} and  \ref{fig:cosine2}. Specifically, we randomly select the ``tulips'' and ``cloud'' classes from CIFAR-100 to compute cosine similarities for positive (same class) and negative pairs (different classes). Observations from these plots reveal that the CLCE embeddings demonstrate superior separation between classes and less overlap between positive and negative samples compared to CE.



\begin{figure}[h]
\vspace{-2mm}
\centering
\includegraphics[width=1\linewidth]{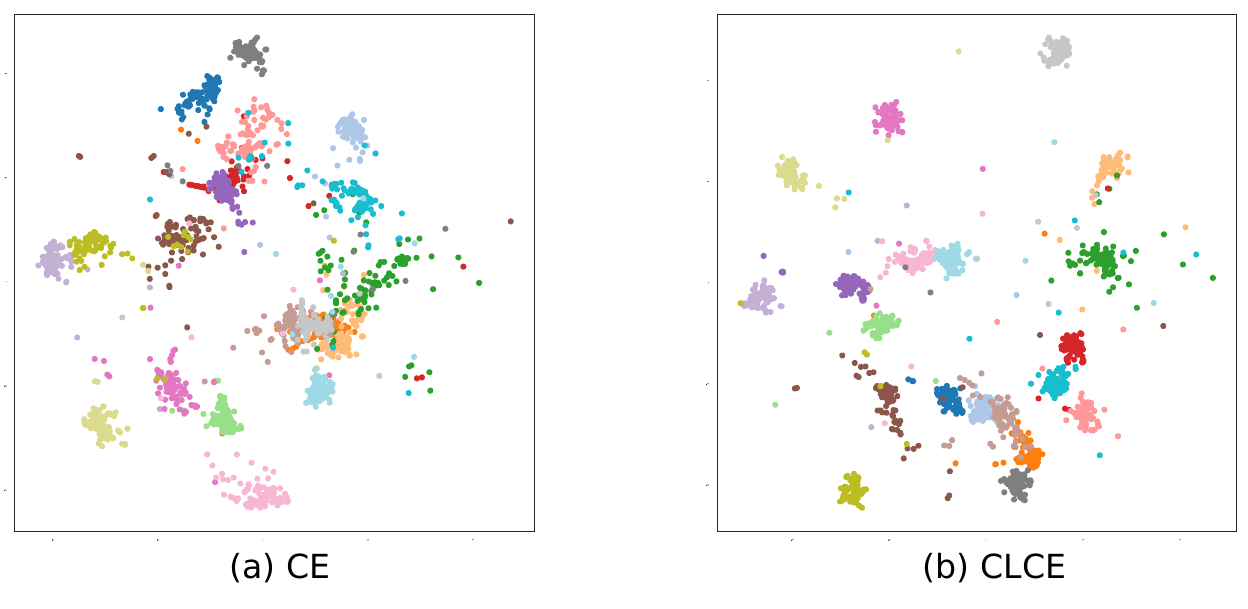}
\caption{\looseness -1 Embedding Space Visualization for CE vs. CLCE, over twenty CIFAR-100 test set classes using t-SNE. Each dot represents a sample, with distinct colors indicating different label classes.}
\label{fig:tsne}
\vspace{-2mm}
\end{figure}

In Fig.~\ref{fig:tsne}, the t-SNE visualization of the embedding space for CE and CLCE across twenty CIFAR-100 classes. The CE embeddings (Fig.~\ref{fig:tsne}a) display instances where the same class nodes are relatively closely packed but also reveal many outliers. This suggests a reduced discriminative capability. On the contrary, CLCE embeddings (Fig.~\ref{fig:tsne}b) display more separated and compact class clusters, suggesting improved discriminative capabilities.

Formally, we calculate the quantitative Isotropy Score (IS)~\citep{mu2018allbutthetop}, which is defined as follows:

\begin{equation}
IS(\mathcal{V}) = \frac{max_{c\subset C} \sum_{v\subset V} \exp{(C^T V)}}{min_{c\subset C} \sum_{v\subset V} \exp{(C^T V)}}
    \label{eq:is}
\end{equation}

where \(V\) is a set of vectors, \(C\) is the set of all possible unit vectors (i.e., any \(c\) so that \(||c|| = 1\)) in the embedding space. In practice, \(C\) is approximated by the eigenvector set of \(V^T V\) (\(V\) are the stacked embeddings of \(v\)). The larger the IS value, the more isotropic an embedding space is (i.e., a perfectly isotropic space obtains an IS score of 1).

\begin{table}
\centering
\resizebox{8.2cm}{!}{\begin{tabular}{@{}lccc@{}}
\toprule
Model        & \multicolumn{1}{l}{iNaturalist2017} & \multicolumn{1}{l}{Imagenet-1k} & \multicolumn{1}{l}{Places365} \\ \midrule
BEiT3-CE     & 0.32                              & 0.27                          & 0.34                        \\
BEiT3-CLCE & 0.98                              & 0.92                          & 0.93                        \\ \bottomrule
\end{tabular}}
\caption{Comparison of Isotropy Score across three datasets for BEiT-3-CE and BEiT-3-CLCE. A higher value is better. A higher Isotropy Score indicates better isotropy and generalizability.}
\label{tab:IS}
\end{table}

Tab.~\ref{tab:IS} demonstrates that the IS score for BEiT-3-CLCE is superior to that of BEiT-3-CE, confirming that CLCE produces a more isotropic semantic space. The BEiT-3-CE embeddings are more anisotropic, implying that BEiT-3-CLCE embeddings more distinctly separate the different classes.

\looseness -1 These observations indicate that the proposed CLCE approach restructures the embedding space to enhance class distinction, addressing the generalization limitation of the CE. This enhancement is particularly effective for few-shot scenarios, where limited labelled data requires the model to rely more on high-quality, discriminative representations.

\section{Discussion and Conclusion}

\paragraph{Limitations.}
While our CLCE approach advances the state-of-the-art, it still has certain limitations. Firstly, CLCE shows increased performance with larger batch sizes. As Table~\ref{tab:batchsize} illustrates, CLCE surpasses CE in accuracy in few-shot and transfer learning scenarios at a batch size of 64, with further improvements observed at larger batch sizes. Secondly, our approach applies hard negative mining solely to the contrastive learning component and not to the CE component. This is due to differing implementations of hard negative mining in each loss. In cross-entropy, hard negatives are identified based on loss values, necessitating a unique strategy that might interfere with the existing sampling process in contrastive learning and potentially cause conflicting outcomes. Additionally, the divergent goals of cross-entropy and contrastive learning, where the former focuses on minimizing the discrepancy between predicted and true distributions and the latter emphasizes embedding similarities, complicate the use of a unified hard negative mining approach.


\paragraph{Conclusion.}

In this work, we proposed a approach for training image models, denoted CLCE. CLCE combines label-aware contrastive learning with hard negative mining and CE, to address the shortcomings of CE and existing contrastive learning methods. 
Our empirical results demonstrate that CLCE consistently outperforms traditional CE and prior contrastive learning approaches, both in few-shot learning and transfer learning settings. Furthermore, CLCE offers an effective solution for researchers and developers who can only access commodity GPU hardware, as CLCE maintains its effectiveness when working with smaller batch sizes that can be loaded onto cheaper GPU cards with less on-board memory. To summarize, our comprehensive investigations and robust empirical evidence compellingly substantiate our methodological decisions, underscoring that CLCE serves as a superior alternative to CE for augmenting the performance of image models for image classification.

\vfill 

\clearpage

\bibliography{mybibfile}

\end{document}